\documentclass[11pt,a4paper]{article}

\usepackage[a4paper,textwidth=120mm,textheight=190mm]{geometry}

\usepackage{mathptmx}
\usepackage{helvet}
\usepackage{courier}

\usepackage{graphicx}

\usepackage[]{units}
\usepackage{url}
\usepackage[pdfpagelabels,plainpages=false,pdfborder={0 0 0}]{hyperref}
\usepackage[all]{hypcap}
\usepackage[]{subfigure}

\begin{document}

\renewcommand{\bottomfraction}{.9}
\renewcommand{\topfraction}{.9}
\graphicspath{{figures/}}
\sloppy

\title{Workspace monitoring and planning\\ for safe mobile manipulation}

\author{Christian Frese, Angelika Zube, and Christian Frey\\
\small{Fraunhofer IOSB, Karlsruhe, Germany}\\
\small{Fraunhofer Institute of Optronics, System Technologies and Image Exploitation}}
\date{}
\maketitle

\begin{abstract}
In order to enable physical human--robot interaction where humans and (mobile) manipulators share their workspace and work together, robots have to be equipped with important capabilities to guarantee human safety. The robots have to recognize possible collisions with the human co-worker and react anticipatorily by adapting their motion to avert dangerous situations while they are executing their task. 

Therefore, methods have been developed that allow to monitor the workspace of mobile manipulators using multiple depth sensors to gather information about the robot environment. This encompasses both 3D information about obstacles in the close robot surroundings and the prediction of obstacle motions in the entire monitored space.
Based on this information, a collision-free robot motion is planned and during the execution the robot continuously reacts to unforeseen dangerous situations by adapting its planned motion, slowing down or stopping. 

For the demonstration of a manufacturing scenario, the developed methods have been implemented on a prototypical mobile manipulator. The algorithms handle both robot platform and manipulator in a uniform manner so that an overall optimization of the path and of the collision avoidance behavior is possible.
By integrating the monitoring, planning, and interaction control components, the task of grasping, placing and delivering objects to humans in a shared workspace is demonstrated.

\end{abstract}

\section{Workspace monitoring}
\label{sec:monitoring}
Monitoring the robot's environment by sensors in order to detect humans and other static or dynamic objects in the robot workspace is a prerequisite for shared human--robot workspaces and for a close collaboration between human and robot. The acquired 3D information about obstacles in the robot's environment is needed by motion control and planning algorithms for collision avoidance. Due to the weakly structured human--robot workspace also information about the objects that have to be handled by the robot and the collaborating humans have to be captured by the robot.

With respect to collision avoidance for fixed-base and mobile manipulators, the information about obstacles should be available in 3D and in a preferably large space around robot in order to be aware of possible dangerous situations at an early stage. On the other hand, the sensor data processing has to be real-time capable due to safety reasons. To cope with these contrary requirements two approaches are combined: \begin{enumerate}
\item A 3D obstacle representation is computed in the close robot surroundings only~\cite{Fetzner14a}. This 3D obstacle model describes the current situation. 
\item In the complete robot workspace, dynamic obstacles are tracked and their motion is predicted in order to predict possible future collisions~\cite{Robotik14}. Obstacle tracking can be performed in a $2\nicefrac{1}{2}$D grid representation, which enables multi-sensor fusion with real-time performance.
\end{enumerate}

\subsection{Obstacle perception using a 3D environment representation}
\label{sec:octree}
For gathering 3D information, depth sensors with different measurement principles are available: time-of-flight sensors (e.g., laser scanner, ToF camera), triangulation based sensors (e.g., Kinect for Xbox 360), or stereo cameras. Several existing approaches concentrate on the usage of one type of depth sensor (e.g., \cite{Flacco12}). But all these sensors can be described by the well-known ray based depth sensor model with a set of rays beginning in the sensor origin. The sensor measures the distances to the points where the rays hit the first object. The measurements of all these different depth sensors can be transformed into point clouds, so that the further data processing becomes independent of the used measurement principle. Exploiting this fact, the developed monitoring algorithm becomes generally applicable to different types of depth sensors.

The sensors can be installed in the workspace or mounted on the robot. Many monitoring approaches deal with sensors fixed in the workspace \cite{Rybski12,Wang12}. But especially in the case of mobile robots, onboard sensors allow to cover the relevant part of large workspaces with only few sensors. So, the developed monitoring algorithm is designed for handling data from different kinds of depth sensors that may be both installed in the workspace and on the robot.

\begin{figure}
  \includegraphics{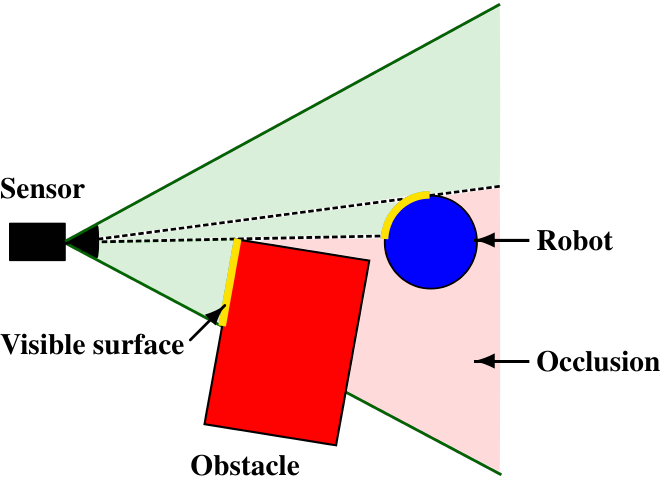}
  \caption{Example of occlusion when an obstacle is located between sensor and robot.}
  \label{fig:Occlusions}
\end{figure}

To ensure safe human--robot interaction, the 3D obstacle representation has to consider occlusions of the robot and of the obstacles. This is particularly important when an object is located between the sensor and the robot (see Fig.~\ref{fig:Occlusions}). Then, considering only the measured obstacle points would lead to an underestimate of the robot--obstacle distance and of the collision risk.
In order to reduce these occlusions and to enhance the overall observed space, information from multiple depth sensors is fused.

The 3D obstacle model is based on an octree data structure. An octree represents the 3D space by a set of nodes. Each node corresponds to a voxel. If necessary, the node is recursively subdivided into eight sub-voxels (nodes), till the minimum voxel size is reached~\cite{Hornung13,Jackins80}. 
The octree is located in the robot's base frame and is limited to the close robot surroundings that have to be monitored in 3D. 

Processing the data of multiple sensors is performed in two steps: Firstly, the data of each sensor is pre-processed independently, in order to allow for parallel computing. Then in the second step, the information is fused to obtain the final obstacle representation.

During the pre-processing procedure, the sensor data is filtered to distinguish between measurements representing obstacles and measurements representing the robot itself, which might be in the field of view of the sensor \cite{URDFFilter}. The filter relies on a geometric robot model. The filtered sensor data is then converted into an octree-based workspace model of the sensor $i$ at the current time $t$: The obstacle measurements are used to compute the set of octree nodes that contain an obstacle point $P_i(t)$. By means of ray tracing, all nodes that are occupied or occluded by an obstacle $O_i(t)$ and all nodes that are occupied or occluded by the robot $R_i(t)$ are computed. Based on the sensor properties, also the set of nodes in the sensor field of view $V_i(t)$ is known. For this reason, the space that can be assumed to be free due to the measurements acquired by sensor $i$ is given by
\begin{equation}
	F_i (t)= V_i (t) \backslash (O_i (t) \cup R_i (t)) .
\end{equation}

\begin{figure}
  \includegraphics[width=\textwidth]{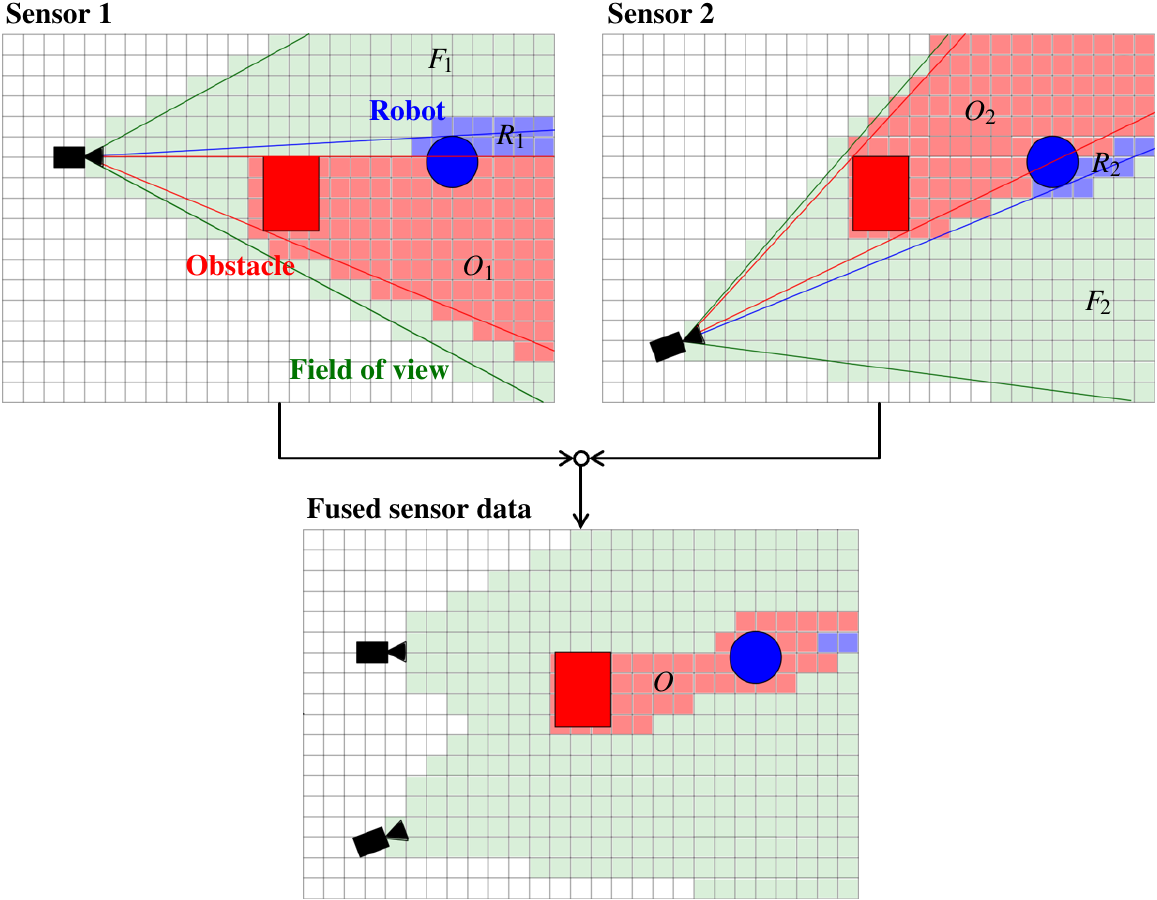}
  \caption{3D workspace monitoring principle with two sensors: fusion of information about occupied and occluded space results in the final obstacle representation $O$.}
  \label{fig:OctreeFusion}
\end{figure}

The information from all available sensors is fused by comparing the obstacle space of one sensor $i$ with the free space of all other sensors:
\begin{equation}
	\widetilde{O}_i (t) = {O}_i (t) \backslash \left(\bigcup_{j,j \neq i} F_j (t) \right) .
\end{equation}
That means $\widetilde{O}_i (t)$ contains only the occluded nodes that are not detected as free by another sensor and the occluded space is reduced as far as the sensor arrangement permits. In the final obstacle representation, the obstacle nodes of all sensors are merged:
\begin{equation}
	O(t)=\bigcup_i \left(P_i (t) \cup \widetilde{O}_i (t) \right) .
\end{equation}

The 3D monitoring principle is illustrated in Fig.~\ref{fig:OctreeFusion} showing an example of two depth sensors. The sensors' fields of view contain the known robot (blue object) and an obstacle (red object). For each sensor, the two upper drawings show the nodes occupied or occluded by the obstacle (red), the nodes occupied or occluded by the robot (blue), and the free nodes (green). The red cells in the lower picture represent the resulting obstacle nodes in the final octree. The occluded space is reduced considerably thanks to the multi-sensor fusion.

Fig.~\ref{fig:OmniRobOctree} shows an example of a 3D obstacle representation obtained from real sensor data on a mobile manipulator. For better understanding, the occupied cells are visualized as red cubes and the occluded cells as rose cubes. The obstacle represents a human standing next to the robot. The human's arm is located between the manipulator and two sensors that are mounted on the platform (see Fig.~\ref{fig:setup}) so that parts of the human in the vicinity of the robot are not visible to the sensors. Considering the occluded space prevents an underestimate of the robot--obstacle distance.

\subsection{Obstacle tracking and prediction}
\label{sec:tracking}
The various object tracking methods presented in the literature can be categorized, among other criteria, in 2D and 3D approaches, and in methods for general objects or specific ones, e.g., vehicles or pedestrians \cite{Lee06a,Moosmann13a,Schulz03a}.

In the context of robot workspace monitoring, it is important to reliably detect moving objects of any kind: humans, trolleys, forklifts, etc. Moreover, multiple sensors are usually required to achieve the desired spatial coverage. These can be homogeneous sensors with complementary fields of view, as usually employed in the literature \cite{Chen13a}, or heterogeneous sensors with considerably different range, resolution, or even sensing principle. In the latter case, particular attention must be paid to the association step in order to achieve the desired robustness for objects moving from one sensor's field of view to a different sensor's field of view.

To enable safe human--robot collaboration, an obstacle tracking approach has been developed which is based on a generic object model and supports multiple heterogeneous sensors, e.g., 2D linescan lasers, 3D laserscanners, and depth cameras. For multi-sensor fusion, the acquired data is mapped into a $2\nicefrac{1}{2}$D grid representation, in which both 2D and 3D sensor data can be integrated \cite{Robotik14}. Indoor environments allow a rather simple obstacle detection: after robot point filtering and possibly outlier detection, any measured 3D point between floor and ceiling can be classified as obstacle point and inserted into the grid. Each obstacle grid cell is annotated with features such as the density of the obstacle points detected by each sensor and the height above ground of the obstacle, if available.

\begin{figure}
	\subfigure[]{\includegraphics[height=56.5mm,trim=80 100 180 100,clip]{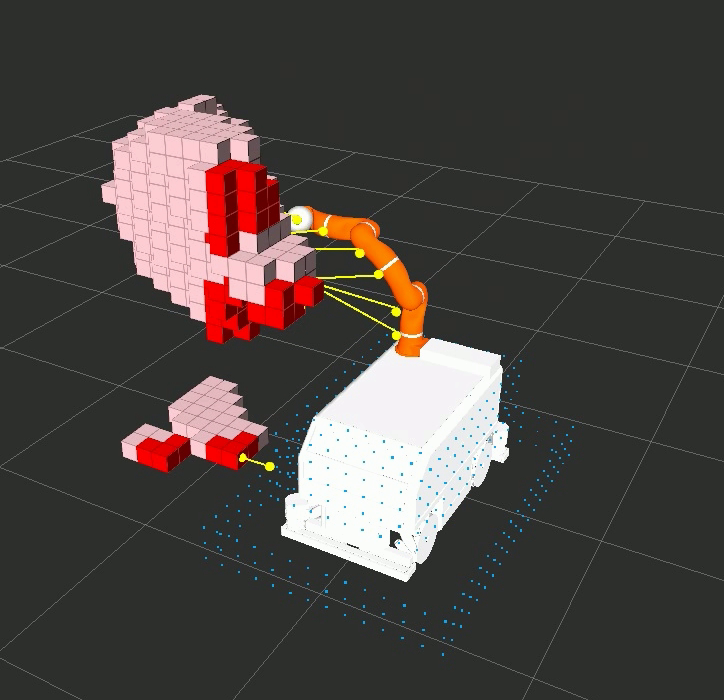}\label{fig:OmniRobOctree}}\hfill
  \subfigure[]{\includegraphics[height=56.5mm]{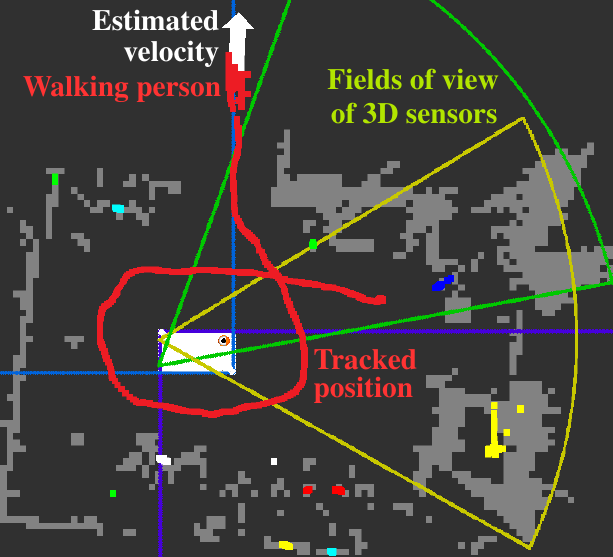}\label{fig:OmniRobTracking}}
  \caption{Examples of monitoring results: \protect\subref{fig:OmniRobOctree} 3D obstacle representation with occupied obstacle cells (red), occluded obstacle cells (rose), and distances between the obstacle and the robot (yellow); \protect\subref{fig:OmniRobTracking}~continuous tracking of a walking person using heterogeneous 2D and 3D sensors.}
  \label{fig:monitoring}
\end{figure}

Object hypotheses $h$ are obtained by clustering connected components of adjacent obstacle cells in the grid. They are associated to existing tracks $o$ by means of a distance function $d_\mathrm{A}(h,o)$ which rates the spatial distance between the estimated centroid position of $h$ and the predicted position of $o$ for the time of the measurement as well as the dissimilarity of the computed features. For the latter, the heterogeneity of the sensors has to be taken into account. For instance, the point densities are compared per sensor, and the height above ground is included in the distance function only if both $h$ and $o$ have been measured by 3D sensors. This approach allows to improve the association quality by incorporating all relevant features available from the sensor data and at the same time to continuously track objects moving between the fields of view of multiple, possibly heterogeneous sensors (Fig.~\ref{fig:OmniRobTracking}).

Kalman filtering based on a linear motion model is applied to estimate the state vector of each obstacle which contains its position and velocity projected onto the ground plane. Based on the estimated velocity and the state covariance, moving obstacles can be detected and their future trajectories can be predicted. The area that is likely to be occupied in the future by the moving obstacle is enlarged depending on the uncertainties estimated by the Kalman filter.

\subsection{Collision avoidance}
\label{sec:CollisionAvoidance}
The information about obstacles detected by the monitoring algorithms is used for a basic collision avoidance strategy. Based on the known geometric robot model, the current minimum distance between the robot and the 3D obstacle representation (see Section~\ref{sec:octree}) is computed. The future minimum distance between the robot and the dynamic obstacles within a certain prediction interval is estimated taking the planned robot motion and the predicted obstacle motion from Section~\ref{sec:tracking} into account. According to the minimum robot--obstacle distance the robot slows down or even stops. In the vicinity of humans this behavior is necessary to ensure safety but also to enhance the comfort level in human--robot cooperation. Additionally to this basic collision avoidance strategy, safety and the task execution velocity may be improved by collision-free motion planning as discussed in Section~\ref{sec:Planning}.

So far, all objects except the robot itself are interpreted as obstacles. But in typical applications as, e.g., pick and place tasks, the classification of what is an obstacle or another kind of object or even which measurement is belonging to the robot may change over time. Generally, all objects have to be considered as obstacles to avoid undesired collisions. But when one object that was an obstacle till now has to be grasped, it should no longer be interpreted as obstacle in order to allow the robot to move close to the object and even to touch it. When the object is grasped by the robot, it has to be (temporary) interpreted as part of the robot, as not only collisions between the robot and its environment but also collisions between the grasped object and the environment have to be avoided.
Therefore, the robot models used for filtering the sensor data and for distance computation are adapted at runtime by adding and removing objects that change their purpose (obstacle, object to be handled, robot part, etc.) depending on the task execution state.

\section{Path planning and adaptation}
\label{sec:Planning}
Robot path planning provides the flexibility required for task accomplishment in dynamic workspaces shared with humans. For typical tasks like pick, place, and handover, the goals are specified as Cartesian end-effector poses. Because the pick/place goal positions are not known beforehand and the scene may change at any time due to dynamic obstacles, this is a typical single-query planning scenario. Moreover, obstacle motions may invalidate the planned path during its execution and require an appropriate reaction, e.g., stopping the robot or adapting the path online.

\subsection{Path planning to Cartesian goal poses}
\label{sec:PathPlanning}
The computational complexity of path planning depends on the degrees of freedom (DoF) of the robot. A mobile platform has 3 degrees of freedom: planar translation in 2 coordinate directions and planar rotation. For these robots, it is feasible to compute optimal paths in a discretized configuration space, using, e.g., A* graph search \cite{Dolgov10a}. Another possibility is the construction of a \mbox{state $\times$ time} lattice based on precomputed motion primitives. These search-based planners allow to consider dynamic obstacles, non-holonomic kinematics, and sequences of multiple goal points~\cite{Petereit13c}.

For robots with more degrees of freedom, optimal planning is no longer feasible under online requirements. Robot arms usually have at least 6 DoF, and mobile manipulators have about 10 DoF. For these robots, sampling-based planning algorithms can be applied~\cite{Kuffner00a}.

\begin{figure}
  \includegraphics{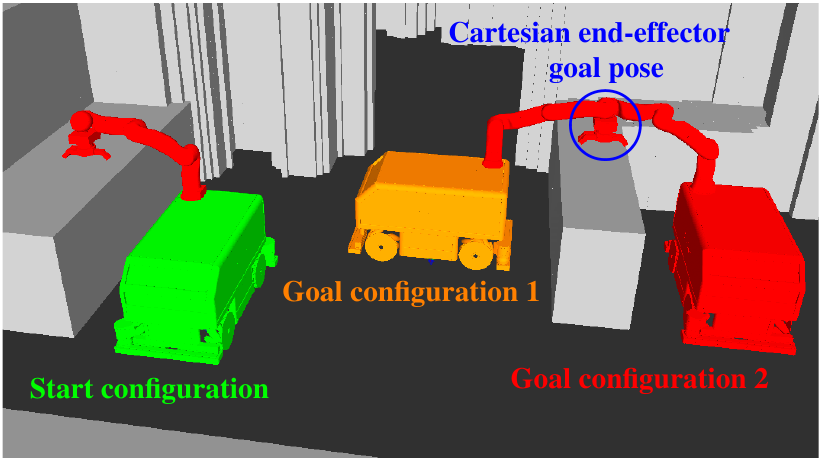}
  \caption{Different goal configurations corresponding to the same Cartesian end-effector goal pose may require paths of considerably different lengths from the start configuration.}
  \label{fig:CartesianGoalExample}
\end{figure}

The usual approach to solve such path planning problems consists of the following two stages:\begin{enumerate}
\item Computation of an inverse kinematics solution, i.e., a collision-free joint configuration in which the robot end-effector is located at the desired Cartesian goal pose.
\item Sampling-based path planning from start to goal in configuration space.
\end{enumerate}
\noindent
This approach is well suited for 6 DoF industrial robots which have a unique inverse kinematics in most cases. For redundant robots having more degrees of freedom than the 6-dimensional Cartesian workspace pose, an infinite number of configurations corresponds to the same end-effector pose. In this case, the two-stage approach is inherently suboptimal: as the goal configuration is chosen disregarding the start position, the resulting paths may become considerably longer than necessary. Fig.~\ref{fig:CartesianGoalExample} illustrates this effect by means of a 10~DoF mobile manipulator (7~DoF lightweight arm plus 3~DoF for position and orientation of the mobile platform in the ground plane). Goal configuration 1 (depicted orange) is much closer to the start configuration (green) than goal configuration 2 (red). The chosen goal configuration might even be unreachable from the start position depending on the obstacle constellation and the resulting configuration space connectivity.

The suboptimal robot
behavior resulting from the two-stage planner may also
be unexpected for humans and may thus result in dangerous
situations, e.g., when the mobile platform starts to
rotate in front of the table even if a motion of the manipulator
would be sufficient to accomplish the task.

Therefore, single-stage planning methods have been proposed
in order to exploit the optimization potential of
redundant robot kinematics \cite{Bertram06a,isr16,Vahrenkamp12a,Weghe07a}. By considering the start configuration
and the lengths of possible collision-free paths,
the goal configuration can be optimized to achieve shorter path lengths and reduced task cycle times.

\begin{figure}[tb]
  \includegraphics{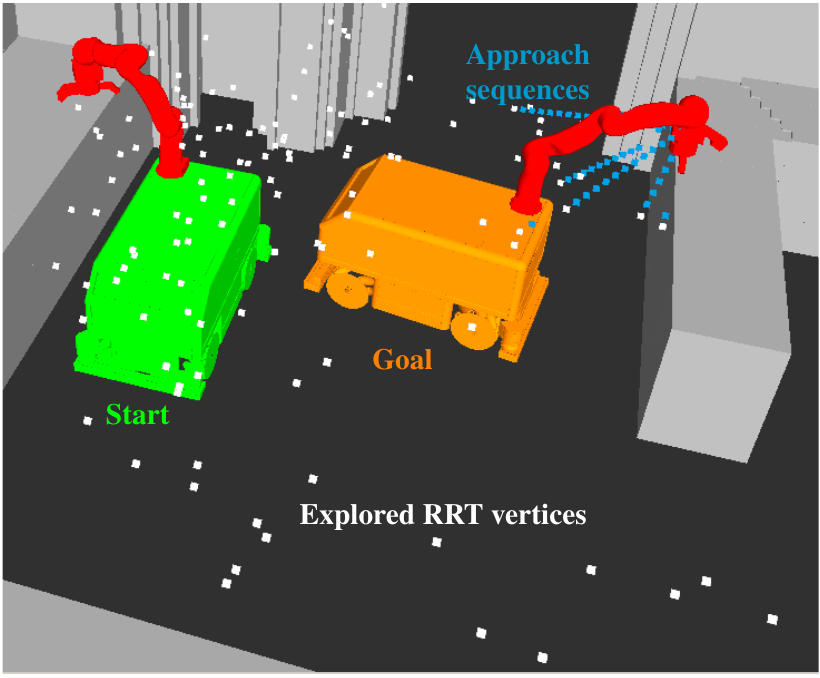}
  \caption{Visualization of goal approach attempts and of collision-free RRT vertices found during exploration. Cartesian end-effector positions of the vertex configurations are shown.}
  \label{fig:CartesianRRTvertices}
\end{figure}

For instance, the well-known rapidly-exploring random tree algorithm \cite{Kuffner00a} can be modified to accept a Cartesian goal pose $\mathbf{p}_\mathrm{goal}$ instead of a goal configuration $\mathbf{q}_\mathrm{goal}\in{Q}$. The distance between an arbitrary configuration $\mathbf{q}$ sampled from the configuration space ${Q}$ and the goal configuration is quantified by means of computing the direct kinematics $\mathbf{f}$ of the robot and applying a distance measure $d_\mathrm{C}(\cdot,\cdot)$ in the 6-DoF Cartesian space (3D position plus 3 DoF for orientation in space),
\begin{equation}
d_\mathrm{C}(\mathbf{f}(\mathbf{q}),\mathbf{p}_\mathrm{goal})\,. \label{eq:goaldistance}
\end{equation}
The expansion of the RRT tree proceeds as usual by sampling collision-free configurations and connecting them to their nearest neighbors in the tree using linear paths in the configuration space. However, it is not possible to directly connect the goal to a vertex of the tree as in the standard RRT algorithm because the goal configuration is not known beforehand. Instead, so-called approach attempts are initiated starting from tree vertices which are close to the goal pose according to (\ref{eq:goaldistance}). An approach is a sequence of vertices connecting the tree vertex to the goal corresponding to a linear path of the end-effector in the Cartesian space (cf.~Fig.~\ref{fig:CartesianRRTvertices}). The configurations $\mathbf{q}$ of the approach vertices are computed from the Cartesian end-effector poses $\mathbf{p}$ by a gradient descent using the Jacobian of the robot kinematics $\mathbf{f}$ subject to constraints such as joint limits \cite{Escande14a}. If an approach is successful, it leads to a goal configuration $\mathbf{q}_\mathrm{goal}$ that satisfies $\mathbf{f}(\mathbf{q}_\mathrm{goal})=\mathbf{p}_\mathrm{goal}$ and is connected to the tree. If a collision occurs in one of the approach configurations, the last collision-free configuration is added to the tree.

An evaluation based on the simulation of a sequence of pick, place, and handover tasks for the mobile manipulator has shown that the described single-stage planner can find shorter paths compared to a two-stage method without increasing the overall computation time \cite{isr16}. Thus it is possible to exploit the kinematic redundancy of the robot to choose an appropriate collision-free goal configuration from the infinite manifold of inverse kinematics solutions and at the same time to reduce the path length.

\subsection{Path smoothing}
\begin{figure}[tb]
  \subfigure[]{\includegraphics[]{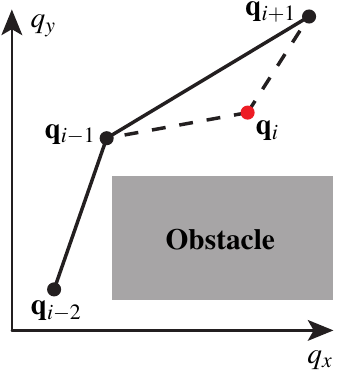}\label{fig:SmoothingRemoveVertices}}\hfill
  \subfigure[]{\includegraphics[]{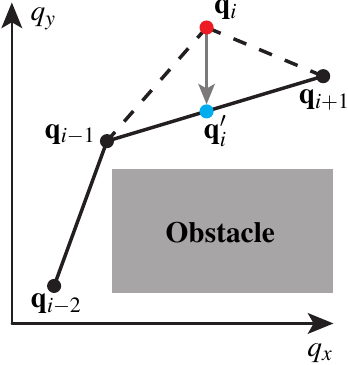}\label{fig:SmoothingIndividualJoints}}\hfill
  \subfigure[]{\includegraphics[]{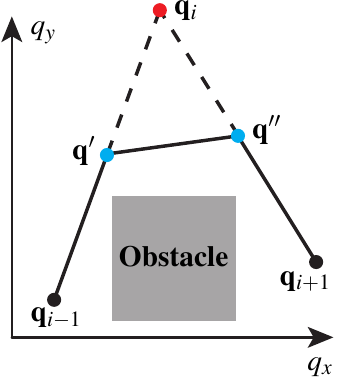}\label{fig:SmoothingCenterConnection}}
  \caption{Path smoothing techniques: \protect\subref{fig:SmoothingRemoveVertices}~removing the unneeded vertex $\mathbf{q}_i$, \protect\subref{fig:SmoothingIndividualJoints}~changing extremum values of individual joints (note that $\mathbf{q}_{i-1}$, $\mathbf{q}_i'$, and $\mathbf{q}_{i+1}$ do not generally constitute a straight line if more than two dimensions are considered), \protect\subref{fig:SmoothingCenterConnection}~adding the connection from $\mathbf{q}'$ to $\mathbf{q}''$ in order to avoid the detour via $\mathbf{q}_i$.}
  \label{fig:Smoothing}
\end{figure}

The first collision-free path obtained by connecting randomly sampled configurations is usually far from the optimal solution and may contain large detours. Two methods to alleviate the suboptimality are\begin{enumerate}
\item path optimization within the sampling-based planner by considering multiple alternative connections for newly sampled configurations and by continuing the path improvement for a certain time after the first solution has been found---as performed, e.g., in the RRT* algorithm \cite{Karaman11a,isr16}, and
\item heuristic path smoothing as a post-processing step.
\end{enumerate}
Heuristic post-processing of the path can be carried out rather fast and may be beneficial even in combination with optimizing sampling-based planners such as RRT*. The path computed by a sampling-based planner is usually represented by a sequence of vertices corresponding to configurations $\mathbf{q}_i$ of the robot which are connected by collision-free linear paths in the configuration space. A simple but effective smoothing technique is to remove unnecessary vertices from the sequence. Thanks to the triangle inequality, removal of one or more vertices is ensured to decrease the overall path length (Fig.~\ref{fig:SmoothingRemoveVertices}). Of course, it has to be verified that the resulting path is collision-free. As the collision checking is the main computational cost of path smoothing, several heuristic strategies have been proposed in order to minimize the path segments to be checked, including\begin{itemize}
\item iteratively removing a single vertex $\mathbf{q}_i$ by directly connecting its predecessor $\mathbf{q}_{i-1}$ and its successor $\mathbf{q}_{i+1}$,
\item binary interval search to find the largest subsequence of vertices $\mathbf{q}_i,\ldots,\mathbf{q}_{i+k}$ that can be removed, and
\item attempting to directly connect two randomly chosen, non-adjacent vertices $\mathbf{q}_i$, $\mathbf{q}_k$, $k>i+1$, of the path.
\end{itemize}
Another smoothing technique does not remove or add vertices, but modifies the configuration of a certain joint in a path vertex. A joint $j$ showing a local extremum $q_{i,j}$ at vertex $\mathbf{q}_i$ with respect to the preceding value $q_{i-1,j}$ and the succeeding value $q_{i+1,j}$ of the joint is set to the intermediate value corresponding to a straight-line motion from $q_{i-1,j}$ to $q_{i+1,j}$ in configuration space---provided that the resulting path segment is collision-free (Fig.~\ref{fig:SmoothingIndividualJoints}). In the special case $q_{i-1,j}=q_{i+1,j}$, joint $j$ does not need to move at all. As sampling-based planners choose all joint configurations at random, it is very unlikely that identical values occur in adjacent vertices. So unnecessary motions result which suggest to a human observer the impression of a purposeless robot behavior. For instance, a mobile manipulator may move its arm back and forth while driving, although it would suffice to move the mobile platform. Such apparently erratic behavior can be substantially reduced by the described path smoothing technique.

Moreover, the joint motions can be interleaved as much as possible in order to reduce the total path execution time. For each path segment, the joint requiring the longest motion time is identified, given the maximum attainable joint velocities. Then it is attempted to shift part of the joint motion to the adjacent path segments if this reduces the required motion time.

In some cases, it can also be beneficial to add new vertices to the path. For adjacent segments of a certain minimum length, it is tested whether the connection of their centers is collision-free (Fig.~\ref{fig:SmoothingCenterConnection}). This procedure is applied recursively. The modified path is again guaranteed to be shorter as the triangle inequality applied to $\mathbf{q}''\mathbf{q}'\mathbf{q}_i$ yields $\overline{\mathbf{q}'\mathbf{q}''}<\overline{\mathbf{q}'\mathbf{q}_i}+\overline{\mathbf{q}_i\mathbf{q}''}$.

\subsection{Path adaptation by elastic bands}
\label{sec:ElasticBands}
In dynamic environments, the planned path may be invalidated during its execution due to obstacles moving unexpectedly into the robot's path. In shared human--robot workspaces, humans may step or grasp into the robot's path at any time. Besides stopping the robot before hitting obstacles as described in Section~\ref{sec:CollisionAvoidance}, the robot may also adapt its plan and continue to move towards its goal. The following principles for plan adaptation can be distinguished:\begin{enumerate}
\item Discarding the invalidated plan and replanning from the current robot state to the goal. Using efficient search-based planners, replanning is possible in real-time for low-dimensional configuration spaces, e.g., for mobile robots \cite{Petereit13c}. The advantage of the replanning principle is that the optimal solution given the current robot state and the current obstacle constellation can be found. Disadvantages include potential oscillations between alternative paths having similar costs and a robot behavior which may be unexpected to humans due to sudden plan changes.
\item Iteratively adapting the planned path to environment changes. Such methods may have lower computational complexity so that real-time path adaptation is feasible for higher-dimensional configuration spaces, e.g., for mobile manipulators~\cite{hfr14}. Another advantage is that the incremental path adaptation creates a behavior which is more predictable for humans. On the other hand, the adapted plans may be suboptimal or the plan adaptation method may fail to find a valid path even if a collision-free solution exists. For instance, these methods usually do not allow the robot to pass a moving obstacle on the opposite side than planned originally.
\end{enumerate}
A method realizing the second principle is the elastic band framework \cite{Quinlan93a}. It has mostly been applied to mobile robots in a 2D environment, but can also handle fixed-base and mobile manipulators. The adapted motion plan is always ensured to be collision-free so that human safety can be guaranteed. The robot follows a smooth path which is adapted smoothly to changes in the environment without unexpected changes of motion direction that could scare human co-workers.

The method has to be initialized with a valid path which can be obtained as described in Section \ref{sec:PathPlanning}. This initial path is then smoothed and adapted to dynamic obstacles observed by the robot's sensors.

\begin{figure}[tb]
  \includegraphics[width=\textwidth]{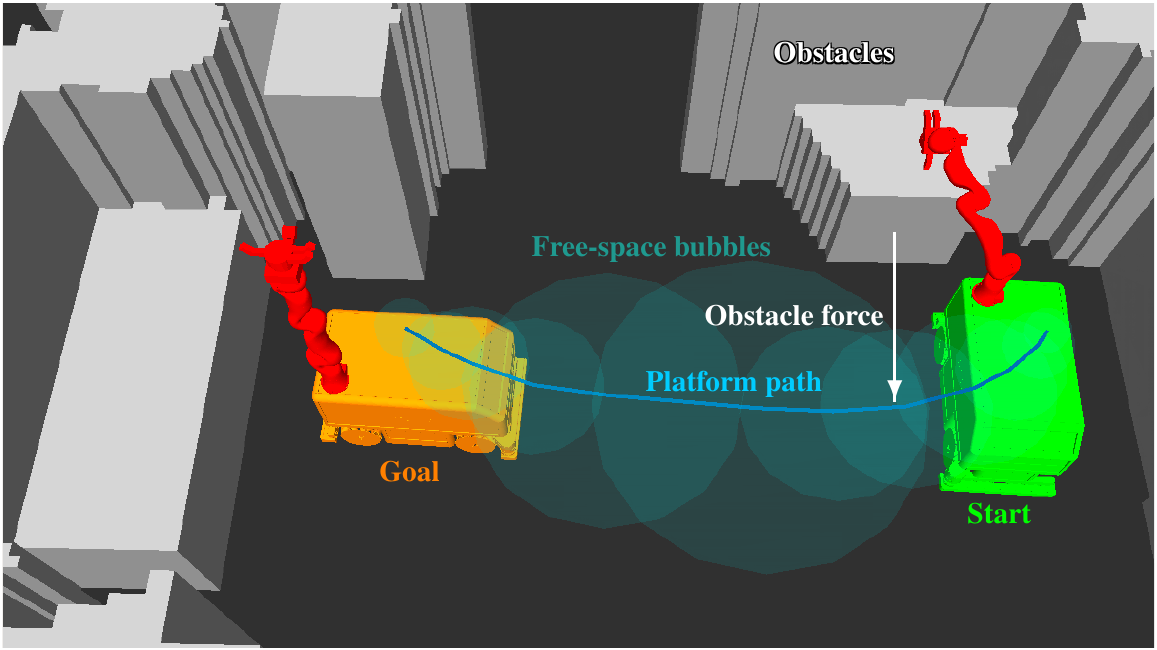}
  \caption{Path adaptation for a mobile manipulator using elastic bands.}
  \label{fig:ElasticBand}
\end{figure}

The elastic band representation of a path consists of a sequence of so-called bubbles $B(\mathbf{q},d)$ which are robot configurations $\mathbf{q}$ annotated with a quantification $d$ of the free space at these configurations, i.e., the distance to the nearest obstacle. Adjacent bubbles always overlap, so that not only the sequence of configurations, but the whole connecting path is ensured to be collision-free (see Fig.~\ref{fig:ElasticBand}). In narrow workspace regions, the bubbles are smaller and thus a greater number of them is required to cover the path.

Artificial forces are computed which push the elastic band towards a smooth curve and repulse it from obstacles. Internal forces model a mutual attraction of neighboring bubbles in order to smooth and shorten the path. For the computation of external forces, the information of the distance and direction to the nearest obstacle is used. For each joint, the algorithm tests whether a motion in any direction increases the distance to the nearest obstacle and chooses the repulsing force accordingly. In this way, the obstacle information is transferred from the 3D workspace to the higher-dimensional configuration space of the robot. It is beneficial to compute the obstacle distances separately for the individual links of the robot, at least for platform and manipulator. Otherwise the motion of the upper manipulator joints has no effect on the computed obstacle distance in many cases, so that no valid force can be obtained.

In more detail, a bubble around a joint configuration $\mathbf{q}\in Q$ can be characterized as follows. If the $j$th revolute joint rotates by $\Delta q_j$, any part of the robot may move at most $r_j(\mathbf{q})\,|\Delta q_j|$ in Cartesian workspace, where $r_j(\mathbf{q})$ is the radius of a cylinder which is aligned to the rotation axis of the joint and contains all subsequent robot links in their considered configuration \cite{Brock}. For a planar translation of a mobile robot by $(\Delta x,\Delta y)^\mathrm{T}$, the motion distance of any robot part is simply given by the length of the translation, $\sqrt{(\Delta x)^2+(\Delta y)^2}$. Altogether, a bubble can be defined as\begin{equation}
B(\mathbf{q},d) := \Bigg\{ \mathbf{q}+\Delta\mathbf{q}\; \Bigg|\; \sum_{j:\;q_j\ \mathrm{revolute}} r_j(\mathbf{q})\,|\Delta q_j| + \sqrt{\sum_{j;\;q_j\ \mathrm{translational}}(\Delta q_j)^2} \,<\, d \Bigg\},
\label{eq:bubbledef}
\end{equation}
where $d$ denotes a lower bound of the clearance at configuration $\mathbf{q}$, i.e., of the minimum Cartesian distance between any part of the robot and any obstacle in the environment \cite{Quinlan93a}. Because bubbles are convex, two bubbles $B(\mathbf{q}_1,d_1)$ and $B(\mathbf{q}_2,d_2)$ are guaranteed to overlap if any configuration $\mathbf{q}$ along the straight line from $\mathbf{q}_1$ to $\mathbf{q}_2$ in configuration space is contained in both bubbles according to~(\ref{eq:bubbledef}). A good candidate for checking this condition is the configuration\begin{equation}
\frac{d_2}{d_1+d_2}\,\mathbf{q_1}+\frac{d_1}{d_1+d_2}\,\mathbf{q_2}\,.
\end{equation}
If adjacent bubbles do not overlap, a new intermediate bubble is inserted into the elastic band. On the other hand, a bubble $B(\mathbf{q}_i,d_i)$ can be removed from the sequence $(\ldots,B(\mathbf{q}_{i-1},d_{i-1}),B(\mathbf{q}_i,d_i),B(\mathbf{q}_{i+1},d_{i+1}),\ldots)\,$ if $\,B(\mathbf{q}_{i-1},d_{i-1})\,$ and $\,B(\mathbf{q}_{i+1},d_{i+1})$ overlap.

The internal contraction force\begin{equation}
\mathbf{F}_{i,\mathrm{int}}:=\frac{\mathbf{q}_{i-1}-\mathbf{q}_i}{\|\mathbf{q}_{i-1}-\mathbf{q}_i\|}+\frac{\mathbf{q}_{i+1}-\mathbf{q}_i}{\|\mathbf{q}_{i+1}-\mathbf{q}_i\|}
\end{equation}
attracts bubble $B(\mathbf{q}_i,d_i)$ to the adjacent configurations $\mathbf{q}_{i-1}$ and $\mathbf{q}_{i+1}$ within the elastic band so that detours are avoided and a smooth motion results \cite{Quinlan93a}.

The obstacle force $\mathbf{F}_{i,\mathrm{obst}}$ pushes each bubble $B(\mathbf{q}_i,d_i)$ away from the nearest obstacle. To compute the obstacle force, not only the distance bound $d_i$, but also the locations of the nearest obstacle points $\mathbf{o}_{ij}$ and the corresponding robot points $\mathbf{x}_{ij}$ for each robot link $j$ are retrieved from the collision test and distance computation library. The obstacle repulsion force for the translational motion of a mobile robot platform can be directly obtained from these values as $\mathbf{F}_{i,\mathrm{obst,trans}}:=s_i\,(\mathbf{x}_{i,\mathrm{trans}}-\mathbf{o}_{i,\mathrm{trans}})$. Therein, the scaling factor\begin{equation}
s_i:=\left\{ \begin{array}{l@{\quad}l} \left(\frac{d_{\max}-d_i}{d_{\max}}\right)^2 & \mathrm{if}\ \;d_i < d_{\max} \\ 0 & \mathrm{if}\ \;d_i \geq d_{\max}\end{array} \right.
\end{equation}
increases the repulsion for closer obstacles, while obstacles with a distance greater than $d_{\max}$ from the robot no longer cause a repulsion. For each revolute joint $j$, it is tested whether a small motion $\Delta q_j$ increases or decreases the robot--obstacle distance $\|\mathbf{x}_{ij}(\mathbf{q}_i,\Delta q_j)-\mathbf{o}_{ij}\|$, where the robot link position $\mathbf{x}_{ij}(\mathbf{q}_i,\Delta q_j)$ corresponding to the modified configuration is computed by means of the kinematic robot model, while $\mathbf{o}_{ij}$ remains fixed. The force for the revolute joint $j$ is then given by
\begin{equation}
F_{ij,\mathrm{obst}}:=s_i\,\frac{\|\mathbf{x}_{ij}(\mathbf{q}_i,+\Delta q_j)-\mathbf{o}_{ij}\|-\|\mathbf{x}_{ij}(\mathbf{q}_i,-\Delta q_j)-\mathbf{o}_{ij}\|}{2\Delta q_j}\,,\quad \Delta q_j>0\,.
\end{equation}
This method of force computation has the advantage that one distance computation is sufficient per iteration and bubble. An alternative method is to call the distance computation library for each configuration modification $\pm\Delta q_j$, which may yield more accurate results if multiple obstacles are close to the robot at the cost of a considerably increased computational effort.

The total force to be applied to each bubble $B(\mathbf{q}_i,d_i)$ is given by a weighted sum of internal and obstacle forces, $\mathbf{F}_i:=\lambda_\mathrm{int}\,\mathbf{F}_{i,\mathrm{int}}+\lambda_\mathrm{obst}\,\mathbf{F}_{i,\mathrm{obst}}$. In each iteration, the bubble configuration is modified according to $\mathbf{q}_i':=\mathbf{q}_i+\varepsilon_i\,\mathbf{F}_i$, where $\varepsilon_i>0$ is chosen to satisfy $\|\varepsilon_i\,\mathbf{F}_i\|<d_i$ in order to ensure that the modified configuration is located inside the bubble and is thus collision-free. Additionally, force components tangential to the elastic band can be suppressed, which helps to avoid oscillations~\cite{Quinlan93a}.

Thanks to the obstacle force, the clearance of the path is increased to an appropriate minimum robot--obstacle distance. In contrast, sampling-based planners usually do not maximize the clearance, but compute collision-free paths which may get arbitrary close to obstacles.

\begin{figure}[tb]
  \includegraphics[width=\textwidth]{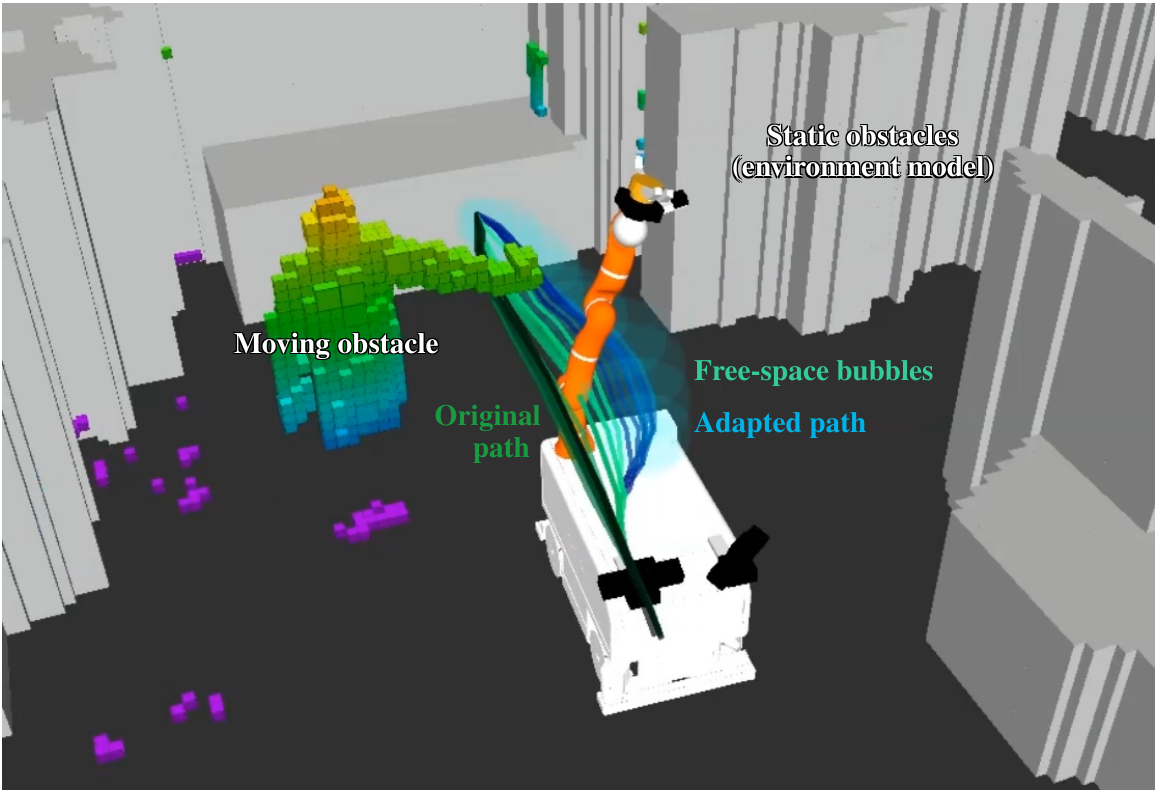}
  \caption{Evasive motion of both mobile platform and manipulator computed by the elastic band method.}
  \label{fig:ElasticControl}
\end{figure}

By applying the artificial forces, the elastic band adapts to dynamic obstacles which are detected by the sensors of the robot and represented in the octree model from Section~\ref{sec:octree}. Smooth evasive motions result which are intuitively understandable to humans. Figure~\ref{fig:ElasticControl} shows an experiment in which the mobile manipulator avoids a human crossing its path. Both mobile platform and manipulator perform a simultaneous, coordinated evasive motion.

If the path is blocked by an obstacle while the robot is moving, this is detected by the elastic band method as the corresponding bubble configurations are no longer collision-free. The robot may then continue to move as long as a sufficient clearance is guaranteed by the bubbles and come to a safe stop before it hits the obstacle. Meanwhile, the elastic band optimization continues so that a collision-free path may be recovered if either a suitable detour is found thanks to the obstacle force or the obstacle is removed from the path.

\section{Physical interaction with humans and objects}
In shared human--robot workspaces, the robot physically interacts with a dynamic environment. This encompasses both interaction with objects (tools, parts, etc.) and interaction with human co-workers.

\subsection{Grasping objects}
\label{sec:grasping}
\begin{figure}[b]
  \subfigure[]{\includegraphics[height=29.5mm]{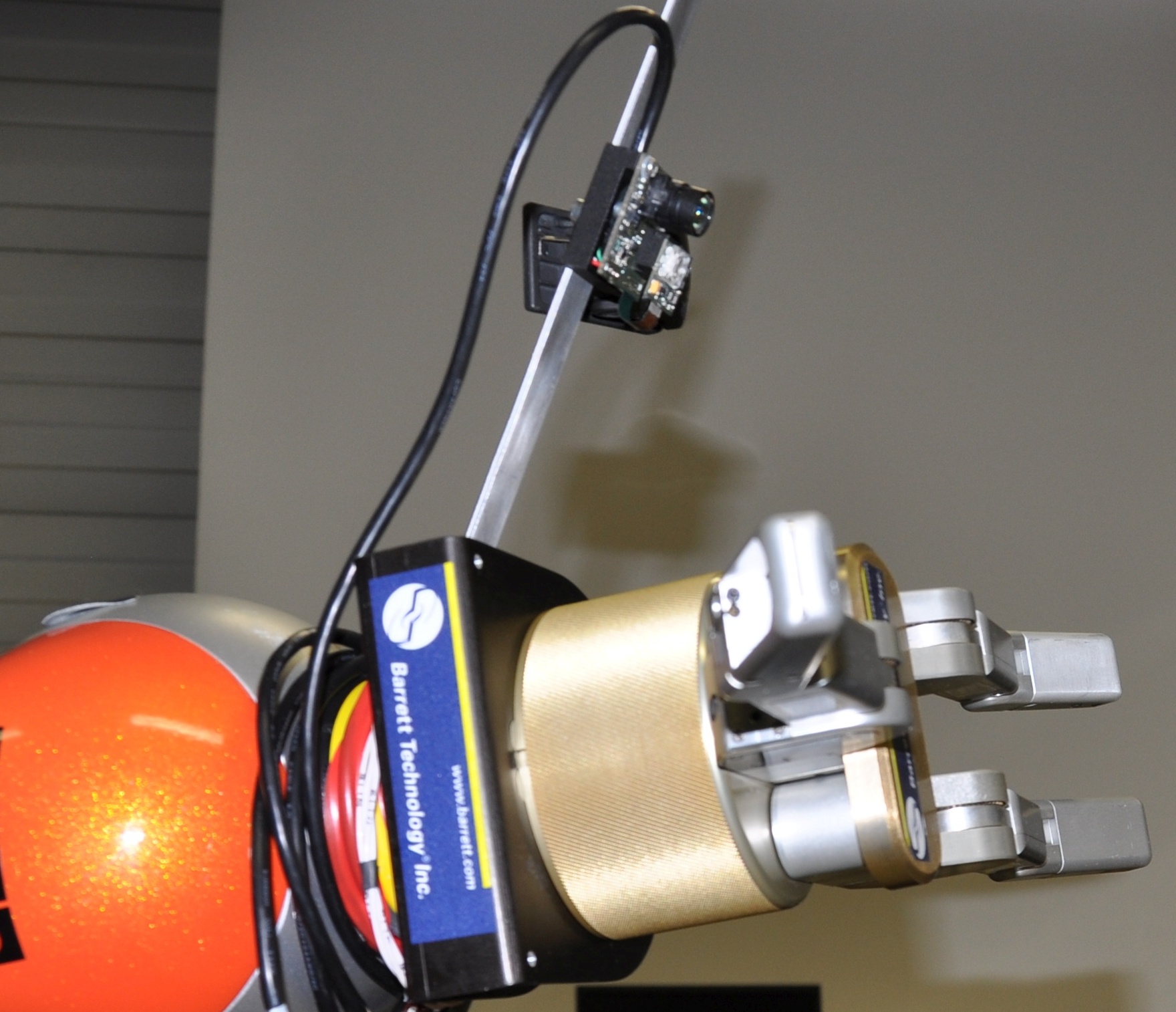}\label{fig:grasping:HandDepthCamera}}\hfill
  \subfigure[]{\includegraphics[height=29.5mm]{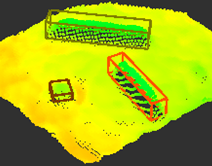}\label{fig:grasping:ObjectBoundingBox}}\hfill
  \subfigure[]{\includegraphics[height=29.5mm]{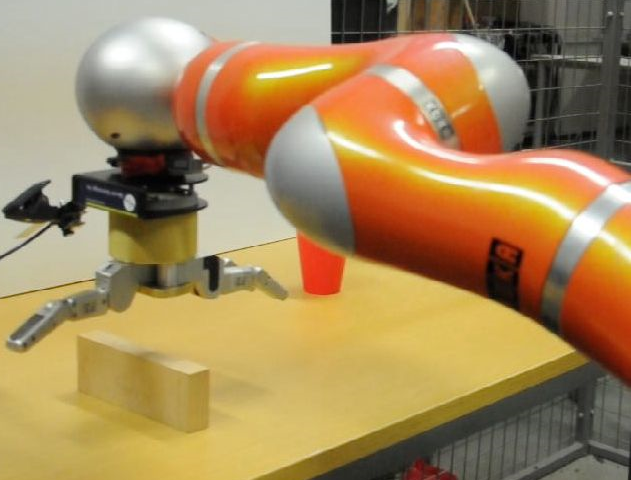}\label{fig:grasping:HandPosition}}
  \caption{Grasping box-shaped objects: \protect\subref{fig:grasping:HandDepthCamera}~depth camera mounted near the robot hand, \protect\subref{fig:grasping:ObjectBoundingBox}~acquired 3D point cloud (height encoded in false colors) and estimated bounding box models of the objects located on the planar surface, \protect\subref{fig:grasping:HandPosition}~positioning the robot hand exactly above the object, with fingers oriented according to the chosen grasping points.}
  \label{fig:grasping}
\end{figure}

An important example of interaction with objects is grasping. When an object is to be grasped, its position and shape are often not known exactly, for instance, because humans may have repositioned the object. Therefore, the robot has to be equipped with appropriate sensors to estimate the object's position and shape. For instance, a depth camera may be mounted near the gripper of the robot, as shown in Fig.~\ref{fig:grasping:HandDepthCamera}.

In the following, a possible processing pipeline for the depth camera data is sketched. A hand--eye calibration is necessary to determine the pose of the camera relative to the tool center point of the robot \cite{Fuchs08a}. Based on the acquired 3D point cloud, planar surfaces like tables are estimated and objects located on top of the surface are segmented. Bounding box models of the objects are estimated (see Fig.~\ref{fig:grasping:ObjectBoundingBox}). Considering the estimated model of the considered object and the limitations imposed by the geometry of the hand, suitable grasping points are selected.

To grasp an object, the robot arm first moves to a position above the table so that it can observe the region of interest in which the object is supposed to be located. Once the object is detected, the hand is positioned exactly above the object and oriented according to the computed grasping points (see Fig.~\ref{fig:grasping:HandPosition}). Then, the robot can move its hand down towards the object without the risk of touching the object accidentally with its fingers. Finally, the fingers are closed to grasp the object.

The grasping behavior control can be implemented by means of a hierarchical state machine. Appropriate reactions to unexpected events and deviations from the planned workflow, as well as other tasks such as placing objects on workbenches or on the mobile platform can also be integrated in this state machine.

\subsection{Physical human--robot interaction}
\label{sec:PhysicalInteraction}
Interaction with humans occurs,
for example, when the robot hands over an object to a human (see Fig.~\ref{fig:demo:handover}). The robot hand releases the object as soon as it detects the force in the robot joints resulting from the human grasping the object. In this context, it is important to distinguish between intended and unintended
interactions, e.g., based on the current interaction situation known from the state machine representation. 
For example, objects handed over to a human again have to be considered as obstacles by the monitoring algorithms.

Intended contacts can furthermore constitute a means to teach the robot or to reposition it. Contact control strategies allow a fast reaction to imminent or detected physical contacts. For example, the kinematic redundancy can be exploited to control the robot joints in a way that allows to reposition the arm in reaction to the contact while maintaining the Cartesian end-effector pose~\cite{Zube16a}.

\section{Prototypes and use cases}
\subsection{Mobile manipulator}
The described monitoring, planning and interaction methods have been integrated on a mobile manipulator and their functionality is demonstrated in a typical use case.

The mobile manipulator consists of the omni-directional mobile platform KUKA OmniRob and the lightweight arm KUKA LWR4. Overall, the mobile manipulator has 10 DoF. 
For grasping tasks, the robot is equipped with a 3-fingered BarrettHand and a PMD CamBoard nano depth camera. The depth camera is mounted close to the tool center point of the manipulator so that it can observe the object to be grasped.

\begin{figure}[tb]
  \subfigure[Mobile manipulator with sensor setup.]{\includegraphics[width=46.45mm]{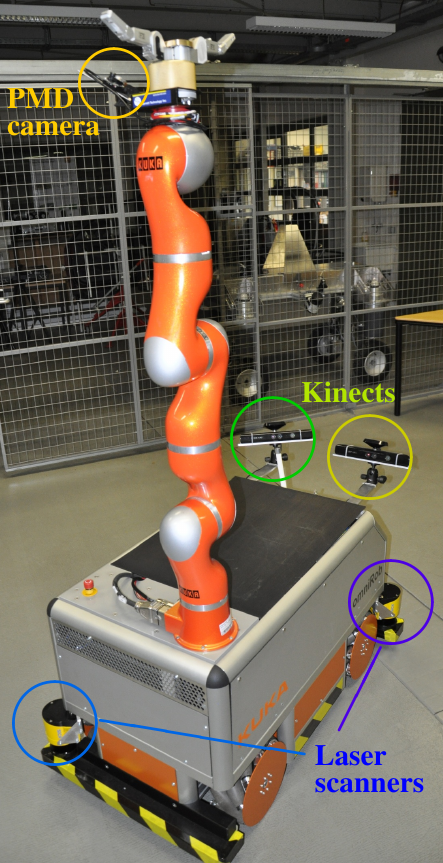}\label{fig:setup}}
	\hfill
	\begin{minipage}[b]{68mm}
	\subfigure[Object hand-over with visualization of the 3D obstacle representation.]{\includegraphics[width=64mm]{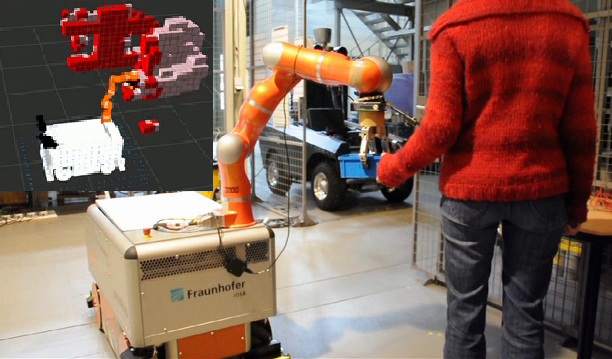}\label{fig:demo:handover}}\vspace{5.5pt}
	\subfigure[Co-worker crossing the robot path with visualization of the obstacle motion prediction.]{\includegraphics[width=64mm]{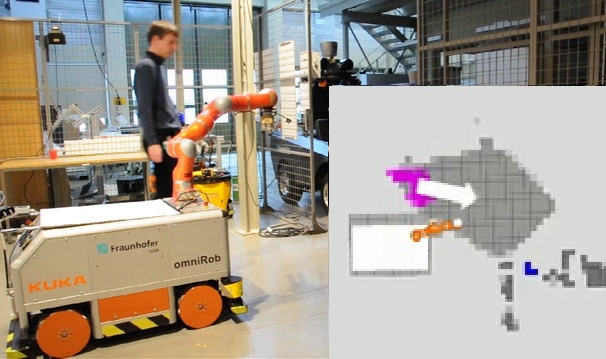}\label{fig:demo:tracking}}
	\end{minipage}
	\caption{Demonstrator setup and snap-shots of the use case.}
	\label{fig:demo}
\end{figure}

For monitoring, onboard sensors are installed on the robot in order to cover the large workspace of the mobile manipulator with only few sensors (see Fig.~\ref{fig:setup}). Two 2D laser scanners (SICK S300) with a $\unit[270]{^{\circ}}$ field of view are mounted at two opposite corners of the platform and monitor a plane around the platform. They detect, e.g., legs of humans next to the platform and objects standing on the floor. Two depth cameras (Microsoft Kinect for Xbox 360) observe the 3D surroundings of the robot arm. They detect, e.g., body and arms of interacting humans. Their placement is chosen in order to achieve a high sensing volume and small occlusions and is determined based on a 3D simulation of the covered space. The extrinsic sensor calibration is performed by registering the acquired 3D point cloud of the manipulator to the robot model, as described in~\cite{Robotik14}.

In the considered use case, the mobile manipulator works in a shared human--robot workspace in an industrial environment. The robot's main task is to transport parts as for example boxes containing screws or tools. The robot has to pick up these parts from a workbench. Their approximate position on the workbench is known, but the exact position and orientation may vary due to the humans interacting with the same parts. Therefore, the objects are detected and localized by means of the grasping methods presented in Section~\ref{sec:grasping}. The robot transports the parts to a second workbench, where it places the parts, or it hands the parts over to a human worker (see Section~\ref{sec:PhysicalInteraction}). Meanwhile, other humans may work at the same workbenches or walk around in the workspace so that they cross the robot path.

Due to the dynamic environment, the robot paths to grasping, placing, and hand-over poses are planned online as described in Section~\ref{sec:PathPlanning}. The planning takes the human co-workers, the furnishings in the workspace and further objects (e.g., on the workbenches) into account. They are detected by the monitoring algorithms (see Section~\ref{sec:monitoring}) that deliver both a 3D-representation of the obstacles in the close robot surroundings and a prediction of the obstacle motions (Fig.~\ref{fig:demo}b,c). During the robot motion execution, the planned path is permanently adapted to the changing environment (see Section~\ref{sec:ElasticBands}). If it is not possible to prevent the robot from a collision by an evasive movement, the robot is slowed down or stopped.

\subsection{Other applications}
Many of the presented monitoring and planning algorithms have also been adapted to other robotic systems, including indoor and outdoor robots. Two examples are:
\begin{itemize}
\item A collaborative assembly station, in which two fixed-base lightweight robot arms support a human worker in performing assembly tasks \cite{Lengenfelder20a}.
\item The autonomous excavator IOSB.BoB \cite{Emter17a}. The excavator is equipped with several 3D laserscanners for environment perception and work\-space monitoring. From the planning perspective, an excavator can be considered as a mobile manipulator, to which sampling-based planning is applicable as described in Section~\ref{sec:Planning}.
\end{itemize}

\section{Conclusions}
In shared human--robot workspaces, it is essential that robots can perceive the dynamic environment and react to moving obstacles in order to avoid imminent collisions. Multi-sensor fusion is necessary in most cases to detect obstacles in the entire relevant space around the robot with sufficient robustness. Motion planning algorithms allow to find collision-free (near-)optimal configuration space paths to goal poses specified online in Cartesian workspace coordinates. Safe and intuitive human--robot interaction can be achieved by combining various sensing, planning, and control methods.

By employing a generic robot model, the described methods are applicable to mobile robots, fixed-base manipulators as well as mobile manipulators, as shown in the presented use case incorporating a 10 DoF mobile manipulator.

\subsection*{Acknowledgement}
The presented research has been supported by the European Commission's 7th Framework Programme as part of the project SAPHARI (Safe and Autonomous Physical Human-Aware Robot Interaction) under grant agreement ICT-287513.

\end{document}